\documentclass[10pt, a4paper]{article}

\usepackage{lrec}
\usepackage{graphicx}
\usepackage{tabularx}
\usepackage{soul}
\usepackage[latin1]{inputenc}
\usepackage[hidelinks]{hyperref}
\usepackage{xstring}

\usepackage{tikz}

\usetikzlibrary{arrows,shapes,positioning,shadows,trees}

\usepackage{listings}
\usepackage{xcolor}

\colorlet{punct}{red!60!black}
\definecolor{background}{HTML}{EEEEEE}
\definecolor{delim}{RGB}{20,105,176}
\colorlet{numb}{magenta!60!black}

\lstdefinelanguage{json}{
    basicstyle=\small\ttfamily,
    numbers=left,
    numberstyle=\scriptsize,
    stepnumber=1,
    numbersep=8pt,
    showstringspaces=false,
    breaklines=true,
    frame=lines,
    captionpos=b,
    backgroundcolor=\color{background},
    literate=
     *{0}{{{\color{numb}0}}}{1}
      {1}{{{\color{numb}1}}}{1}
      {2}{{{\color{numb}2}}}{1}
      {3}{{{\color{numb}3}}}{1}
      {4}{{{\color{numb}4}}}{1}
      {5}{{{\color{numb}5}}}{1}
      {6}{{{\color{numb}6}}}{1}
      {7}{{{\color{numb}7}}}{1}
      {8}{{{\color{numb}8}}}{1}
      {9}{{{\color{numb}9}}}{1}
      {:}{{{\color{punct}{:}}}}{1}
      {,}{{{\color{punct}{,}}}}{1}
      {\{}{{{\color{delim}{\{}}}}{1}
      {\}}{{{\color{delim}{\}}}}}{1}
      {[}{{{\color{delim}{[}}}}{1}
      {]}{{{\color{delim}{]}}}}{1},
}

\usepackage{titlesec}
\titleformat{\section}{\normalfont\large\bf\center}{\thesection.}{1em}{}
\titleformat{\subsection}{\normalfont\SmallTitleFont\bf\raggedright}{\thesubsection.}{1em}{}
\titleformat{\subsubsection}{\normalfont\normalsize\bf\raggedright}{\thesubsubsection.}{1em}{}
\renewcommand\thesection{\arabic{section}}
\renewcommand\thesubsection{\thesection.\arabic{subsection}}
\renewcommand\thesubsubsection{\thesubsection.\arabic{subsubsection}}

\title{A Workflow Manager for Complex NLP and Content Curation Pipelines}

\name{Juli\'{a}n Moreno-Schneider, Peter Bourgonje, Florian Kintzel, Georg Rehm}

\address{Speech and Language Technology Lab, DFKI GmbH \\
         Alt-Moabit 91c, 10557 Berlin, Germany \\
         \{julian.moreno\_schneider, peter.bourgonje, florian.kintzel, georg.rehm\}@dfki.de\\}

\abstract{We present a workflow manager for the flexible creation and customisation of NLP processing pipelines. The workflow manager addresses challenges in interoperability across various different NLP tasks and hardware-based resource usage. Based on the four key principles of \emph{generality}, \emph{flexibility}, \emph{scalability} and \emph{efficiency}, we present the first version of the workflow manager by providing details on its custom definition language, explaining the communication components and the general system architecture and setup. We currently implement the system, which is grounded and motivated by real-world industry use cases in several innovation and transfer projects. 
\\ \newline  
}

\begin{document}

\maketitleabstract

\section{Introduction}
\label{sec:introduction}

The last decades have seen a significant increase of digital data. To allow humans to understand and interact with this data, Natural Language Processing (NLP) tools targeted at specific tasks, such as Named Entity Recognition, Text Summarisation or Question Answering, are under constant development and improvement to be used together with other components in complex application scenarios. While several NLP tasks can be considered far from being solved and others increasingly maturing, one of the next challenges is the combination of different task-specific services based on modern micro-service architectures and service deployment paradigms. 

Chaining tools together by combining their output requires not much more than simple interoperability regarding the annotation format used by the semantic enrichment services and individual NLP services. However, the notion of flexible workflows stretches, beyond annotation formats, to the flexible and efficient orchestration of NLP services. While a multitude of components and services is available, the next step, i.\,e., the management and integration into an infrastructural system, is not straightforward and proves challenging. This is problematic both for technology developers and users, as the whole is greater than the sum of its parts. Developers can add value to their tools by allowing the combination with other components. For users, the benefits of combining annotations obtained from NER with those obtained by coreference resolution, for example, are obvious. There have been several attempts, both commercial and open source, to address interoperability and service orchestration for scenarios that include the processing of document collections, achieving comparatively good results for specific use cases, tasks and domains (see Section~\ref{sec:relatedwork} for an overview).

Recently, new opportunities have been generated by the popularity of containerisation technologies (such as Docker\footnote{\url{https://www.docker.com}}), that enable the deployment of services and tools independently from the environment in which they were developed. While integration benefits from this approach by enabling easy ingestion of services, the methodology comes with several challenges that need to be addressed, including, crucially, container management. This is not just about keeping services alive on different nodes, which can be done using tools such as Kubernetes\footnote{\url{https://kubernetes.io}} or Openshift\footnote{\url{https://www.openshift.com}}. The key challenge remains allowing the organisation and inter-connectivity of services in terms of their functionality, ensuring that they work together in an efficient and coordinated way. 

The work presented in this paper is carried out under the umbrella of the QURATOR project\footnote{\url{https://qurator.ai}}, in which a consortium of ten partners (ranging from research to industry) works on challenges encountered by the industry partners in their own specific sectors. The central use case addressed in the project is that of \emph{content curation} in the areas of, among others, journalism, museum exhibitions and public archives \cite{rehm2020d,rehm2016j}. In QURATOR, we develop a platform that allows users to curate large amounts of heterogeneous multimedia content (including text, image, audio, video). The content is collected, converted, aggregated, summarised and eventually presented in a way that allows the user to produce, for example, an investigative journalism article on a contemporary subject, content for the catalogue of a museum exhibition, or a comprehensive description of the life of a public figure, based on the contents of publicly available archive data on this person. To achieve this, we work with various combinations of different state-of-the-art NLP tools for NER, Sentiment Analysis, Text Summarisation, and several others, which we develop further and integrate into our platform. The interoperability and customisation of workflows, i.\,e., distributed processing pipelines, are a central technical challenge in the development of our platform.

The key contribution of this paper is the presentation of a novel workflow management system aimed at the sector-specific content curation processes mentioned above. Technically, the approach focuses on the management of containerised services and tools. The system design is optimised and aligned with regard to four different dimensions or requirements: (i) \emph{generality}, to work with a diverse range of containerised services and tools, independent of the (programming) language or framework they are written in; (ii) \emph{flexibility}, to allow services or tools -- which may be running on different machines -- to connect with one another in any order (to the extent that this makes sense, semantically); (iii) \emph{scalability}, to allow the inclusion of additional services and tools; and (iv) \emph{efficiency}, by avoiding unnecessary overhead in data storage as well as processing time.

The rest of the paper is structured as follows. Section~\ref{sec:relatedwork} describes approaches similar to ours that support the specification of workflows for processing document collections. Section~\ref{sec:sys_overview} provides an overview of the proposed system and lists requirements regarding the services to be included in workflows. Section~\ref{sec:cwdl} presents the workflow specification language. Section~\ref{sec:architecture} outlines the general architecture and the following subsections provide more detail on individual components.
Finally, Section~\ref{sec:conclusionsfuturework} concludes the article and sketches directions for future work.

\section{Related Work}
\label{sec:relatedwork}

The orchestration and operationalisation of the processing of large amounts of content through a series of tools has been studied and tested in the field of NLP (and others) from many different angles for decades. There is a sizable amount of tools, systems, frameworks and initiatives that address the issue but their off-the-shelf applicability to concrete use cases and heterogeneous sets of services is still an enormous challenges.

One of the most well known industry-driven workflow definition languages is Business Process Model and Notation (BPMN, and its re-definition BPMN~V2.0) \cite{omg2011bpmn}. Many tools support BPMN, some of them open source (Comidor, Processmaker, Activiti, Bonita BPM or Camunda), others commercial (Signavio Workflow Accelerator, Comindware, Flokzu or Bizagi). There are also other business process management systems, not all of which are based on BPMN, such as WorkflowGen\footnote{\url{https://www.workflowgen.com}}, ezFlow\footnote{\url{http://www.ezflow.it}}, Pipefy\footnote{\url{https://www.pipefy.com}}, Avaza\footnote{\url{https://www.avaza.com}} or Proces.io\footnote{\url{http://proces.io}}. Their main disadvantage with regard to our use case is that they primarily aim at modelling actual business processes at companies, including support to represent human-centric tasks (i.\,e., foreseen human interaction tasks). This focus on support deviates from our use case, in which a human user interacts with the content, but not necessarily with other humans.

Another class of relevant software are frameworks for container management, focusing on parallelisation management, scalability and clustering. Examples are Kubernetes, Openshift, Rancher\footnote{\url{https://rancher.com}} and Openstack\footnote{\url{https://www.openstack.org}}. We use Kubernetes for cluster management. However, because this does \emph{not} cover (NLP) task orchestration or address interoperability, with our workflow manager we go beyond the typical Kubernetes use case.

On the other hand, there are numerous frameworks and tool kits that focus more on workflow management and the flexible definition of processing pipelines (and less on the technical, hardware related implementations like Kubernetes, Openshift and Rancher). Examples are Apache Kafka\footnote{\url{https://kafka.apache.org}}, a distributed streaming platform; Apache Commons\footnote{\url{http://commons.apache.org/sandbox/commons-pipeline/}}; Apache NIFI\footnote{\url{https://nifi.apache.org}}; Apache Airflow\footnote{\url{https://airflow.apache.org}}; Kylo\footnote{\url{https://kylo.io}}; and Apache Taverna\footnote{\url{https://taverna.incubator.apache.org}}. With our workflow manager, we attempt to cover these workflow-focused features, but, crucially, combine them with the more technical details for cluster management and scalability.

Specifically targeted at NLP, some popular systems are GATE \cite{Cunningham2011a} and UIMA \cite{UIMA:FERRUCCI:2004}, and, more recently (but covering a narrower range of tasks), SpaCy\footnote{\url{https://spacy.io}}. While the data representation format is based on a standard format for some of these (GATE for example supports exporting data in XML), we attempt to extend beyond this and use the NLP Interchange Format (NIF) \cite{hellmann2013}. Using NIF ensures interoperability for different NLP tasks while at the same time addressing storage and scalability needs. Since NIF is based on RDF triples, the resulting annotations can be included in a triple store to allow for efficient storage and querying. In addition, the above-mentioned systems are designed to run on single systems. Our workflow manager is designed to combine output from different micro-services that address different NLP services, potentially running on different machines. In addition to the above, CLARIN \cite{hinrichs-krauwer-2014-clarin} provides an infrastructure for natural language research data and tools. The focus, however, is on sharing resources and not on building NLP pipelines or workflows. A more exhaustive and complete overview of related work can be found in \cite{elg2020}. 

\section{System Overview}
\label{sec:sys_overview}

The objective of the QURATOR project is to facilitate the execution of complex tasks in the area of content curation. The human experts performing these tasks typically have limited technical skills and are expected to analyse, aggregate, summarise and re-arrange the information contained in the content collections they work with. The Curation Workflow Manager aims to support these users, by allowing them to flexibly and intuitively define just the workflow they need. Ultimately, the aim is to make this as intuitive as using a single call to a single system. The single system will be the Workflow Manager, and the single call will be the request to process the document collection using a specific workflow. The workflow includes all the needed services (i.\,e., which services, such as NER, summarisation, topic modeling, clustering, etc.~to include, and which parameters, such as language or domain, to set). The order of the services, and which can be parallelised, can be specified, as well as which data needs to be stored internally (for immediate processing) or externally. Afterwards, the processed content collection is meant to be presented in a GUI, featuring the relevant data visualisation components, given the original document collection and the result of the individual semantic enrichment processes that have run.

While from a user's perspective, this high level description may sound similar to comparable systems like GATE (Section~\ref{sec:relatedwork}), the following description provides an idea of the intended deployment scale and ambition of the workflow manager. Though developed in the context of the QURATOR project, we plan to implement the workflow manager also in the technical platform architecture developed in the project European Language Grid.\footnote{https://www.european-language-grid.eu}. The main objective of the project ELG is to create the primary platform for Language Technology in Europe \cite{elg2020}. Release 1 alpha of the European Language Grid platform was made available in March 2020 and provides access to more than 150 services including NER, concept identification, dependency parsing, ASR and TTS.

\subsection{Service Requirements}
\label{sec:cwm:taxonomy}

Since we want to allow for the inclusion of as many different services as possible in a workflow, yet at the same time have to ensure that they work together seamlessly, we specified a few core dimensions along which to classify services, to establish whether or not they can be included. First (i), we check whether a service is dockerised or not. Then (ii), we check the execution procedure, i.\,e., is it a fully automated service, or is human intervention or interaction included, or even at the very core (such as, for example, in annotation editors). Furthermore, we check (iii) where the service is located, i.\,e., is it included in the Docker cluster or is it a service hosted externally? Finally (iv), we check how the service is communicating, i.\,e., is it accessible through a REST API or a command-line interface? If a given service is (i) dockerised (or otherwise containerised), (ii) does not need human intervention, (iii) is stored inside our Docker cluster and (iv) has a REST API interface through which it can be accessed, we conclude that the service can be included in our workflow.\footnote{As part of future work we will investigate if and how these core dimensions can be included in the metadata scheme that governs all metadata entries for all services in order to automate this process as much as possible \cite{labropoulou2020l}.}

\section{Curation Workflow Definition Language}
\label{sec:cwdl}

To facilitate the definition of workflows for users with limited technical knowledge (i.\,e., little to no programming experience), we opted for the widely used JSON format to specify workflows, considering that the specification of actual workflows will be carried out using a corresponding graphical user interface.

We specified a JSON-based Curation Workflow Definition Language (CWDL). It currently supports the inclusion of services with REST API access \cite{Richardson2013} (i.\,e., services must be accessible through HTTP calls), and allows users to specify whether these services should be executed in a synchronous or asynchronous way. The execution in a sequential or parallel fashion can also be specified.

A workflow relies on three main components: \emph{controllers}, \emph{tasks} and \emph{templates}. The \emph{controllers} element relates to a service to be included. This element communicates basic identity information (\texttt{controllerName}, \texttt{serviceId}, \texttt{controllerId}), queue information (\texttt{nameInput\{Normal|Priority\}}) and connection information (\texttt{connection}) to the micro-services it is calling. The \texttt{connection} element contains information needed to communicate with the service (via REST API), including \texttt{method}, \texttt{endpoint\_url}, \texttt{parameters}, \texttt{headers} and \texttt{body}. Listing~\ref{cod:controllerexample} shows an example. 

The next element, \emph{task}, sends messages to and from a controller through the messaging control system. The \texttt{taskId} and \texttt{controllerId} fields contain identifying information on the two. Listing~\ref{cod:taskexample} illustrates this using an example.

\begin{lstlisting}[language=json,frame=single,caption=Example of a Controller definition that connects to an external REST API service.,label=cod:controllerexample]
{
 "controllerName": "NER Controller",
 "serviceId": "NER",
 "controllerId": "NERController",
 "queues": {
  "nameInputNormal": "NER_input_normal",
  "nameInputPriority": "NER_input_prio"
 },
 "connection": {
  "connection_type": "restapi",
  "method": "POST",
  "endpoint_url": "http://<host>/path/",
  "parameters": [
   {"name": "language","type": "parameter",
   "default_value": "en","required": true},
   {"name": "models","type": "parameter",
   "default_value": "model_1;model_2","required": true},
   ...],
  "body": {
   "content": "documentContentNIF"
  },
  "headers": [
   {"name": "Accept","type": "header",
    "default_value": "text/turtle","required": true},
   {"name": "Content-Type","type": "header",
    "default_value": "text/turtle","required": true}
  ]
 }
}
\end{lstlisting}

\begin{lstlisting}[language=json,frame=single,caption=Example of a Task definition.,label=cod:taskexample]
{
  "taskName": "NER Task",
  "taskId": "NERTask",
  "controllerId": "NER",
  "component_type": "rabbitmqrestapi"
}
\end{lstlisting}

The third element, \emph{template}, specifies which micro-services are included in the workflow. Basic identification information is specified in \texttt{workflowTemplateId}. The different micro-services included in the template are contained in \texttt{tasks}. Inside this element, the following information is specified:

\begin{enumerate}
\item \texttt{ParallelTask} executes multiple tasks in parallel.
\item \texttt{SequentialTask} executes tasks sequentially.
\item \texttt{split} splits the input information to every output.
\item \texttt{waitcombiner} waits until all connected inputs have finished to combine their results and proceed.
\end{enumerate}
 
Listing \ref{cod:workflowexample} shows an example of the \texttt{template} element.

\begin{lstlisting}[language=json,frame=single,caption=Example of a workflow definition.,label=cod:workflowexample]
{
 "workflowTemplateName": "GLK",
 "workflowTemplateId": "ML_GLK",
 "workflowTemplateDescription": "...",
 "tasks": [
  {
   "order": 1,
   "taskId": "ParallelTask",
   "features":{
    "input": {"component_type": "split"},
    "output": {"component_type": "waitcombiner"},
    "tasks":[
     {"order": 1, "taskId": "NERTask"},
     {"order": 2, "taskId": "GEOTask"},
     ...]
   }
  },
  ...]
}
\end{lstlisting}

We plan to improve this basic scheme and will make it compliant with BPMN V2.0 in its next iteration.

\section{Curation Workflow Manager Architecture}
\label{sec:architecture}

\begin{figure*}[ht]
  \centering
  \includegraphics[width=0.8\linewidth]{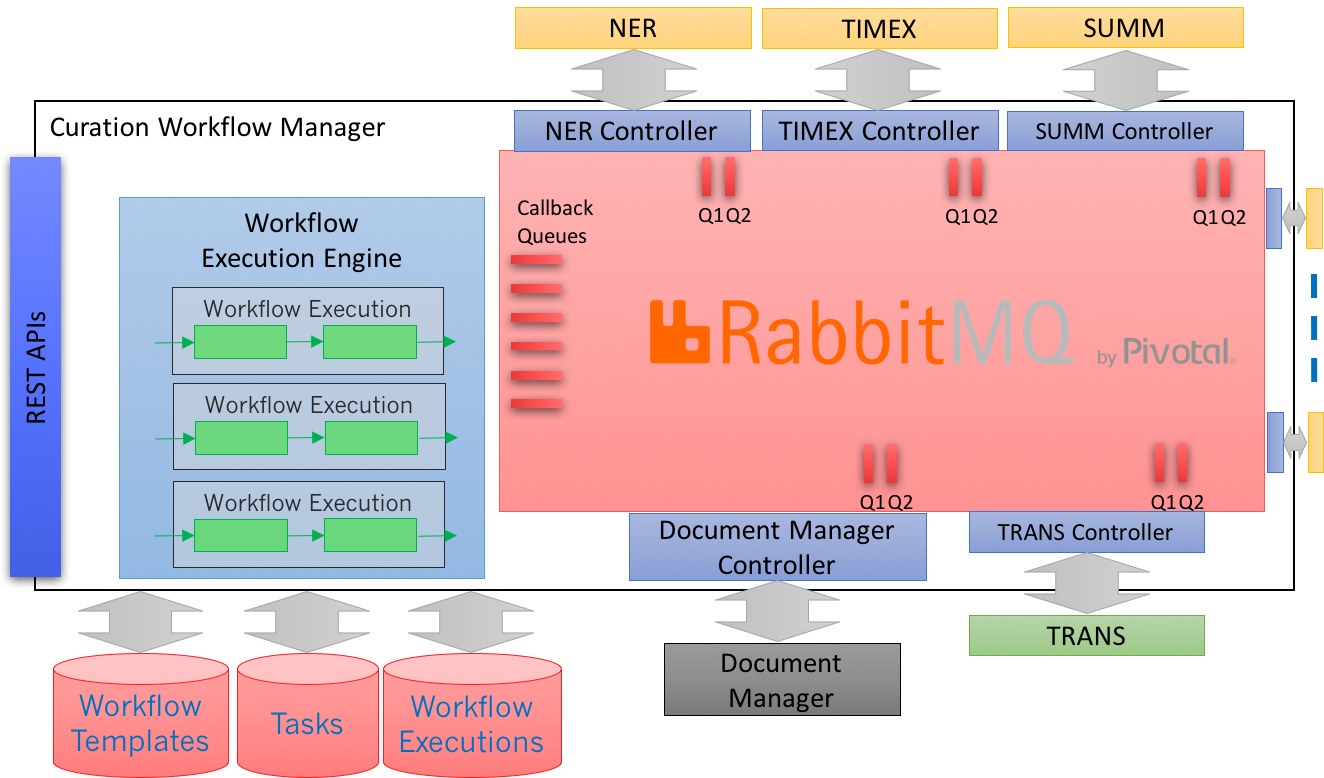}
  \caption{Architecture of the Curation Workflow Manager (CWM)}
  \label{fig:architecture}
\end{figure*}

In Section~\ref{sec:cwdl}, the description of the JSON-based workflow definition language outlines how to instruct the workflow manager to perform complex tasks. In this section, we outline how these task definitions are translated into processes and procedures, by explaining the workflow manager architecture. Our previous work includes a generic workflow manager for curation technologies \cite{rehm2016j,rehm2020d}, and two indicative descriptions of an initial prototype of a workflow manager that we conceptualised based on use cases in the legal domain \cite{rehm2018g,rehm2018f}. Figure~\ref{fig:architecture} illustrates this architecture, its individual components are described in the following subsections.

\subsection{Workflow Execution Engine}

The core component of the workflow manager is the Workflow Execution Engine. This component manages workflows, from their definition to the management of its execution to the final results that are produced. In the CWM a workflow is composed of the three components described in Section~\ref{sec:cwdl}, and a workflow execution. More specifically:

\begin{itemize}
\item A \emph{controller} is a component whose main purpose is to communicate with a service (see Section~\ref{sec:architecture:controllers}).
\item A \emph{task} can be anything that has to do with taking input in a certain format, and producing output. This can be enriching text though NLP components, converting data to a required format for specific other tasks, combining information from different upstream tasks, or deciding which task to perform next, depending on parameters that are either set in the configuration, or that are the outcome of upstream processing.
\item A \emph{template} is an abstract definition of a workflow composed of a combination of tasks. It is, in the literal sense of the word, a preset for a collection of tasks that together form a logical processing pipeline. In the object-oriented programming paradigm, it would be the equivalent of a class, i.\,e., the definition of an object (and the objects would be \emph{tasks}).
\item A \emph{workflow execution} is an instance of a workflow template, i.\,e., a complete workflow created with specific \emph{task} objects. The \emph{workflow execution} would be equivalent to an instantiated object in the object-oriented analogy.
\end{itemize}

\subsection{Controllers}
\label{sec:architecture:controllers}

Every service is required to be accessible through a REST API and must allow both sending and receiving of task-specific messages. Because the services are independently developed, and their behaviour may change with new versions, the way to communicate with them may change as well. We, therefore, introduce the concept of a proxy element between the messaging control system (for which we use RabbitMQ, see Section~\ref{sec:architecture:rabbitmq}) and the service. This proxy element is the \emph{controller}. We attempt to maximise flexibility by updating the \emph{controller} whenever the service changes, so that the rest of the communication chain can remain untouched. 

In the current implementation, the controller connects to RabbitMQ and waits for receiving messages. Whenever a message is received, the controller processes its contents and generates a HTTP request for the corresponding service. Depending on whether or not the service in question executes in a synchronous or asynchronous way, the controller waits for the response, or checks back in to collect it later, and subsequently communicates the result.

\subsection{Communication Module}
\label{sec:architecture:rabbitmq}

The communication module, based on the message control system RabbitMQ, allows the exchange of information between the different workflow components, or with components external to the workflow. As mentioned above, our system requires individual services to be accessible via REST API, and supports both synchronous and asynchronous execution of services.

This communication entails both information relating to tasks to be performed, as well as the result or output of the tasks themselves. We use RabbitMQ, because it allows larger message contents than some of its competitors (Apache Kafka, for example). RabbitMQ handles the communication between the workflow execution engine and the services (through \emph{controllers}). Both the workflow execution engine and the \emph{controllers} send messages to and receive messages from RabbitMQ during the execution of a workflow. The workflow execution engine sends a message to every service (through its proxy element, the \emph{controller}), to execute a processing step. After finishing the processing, the service sends a new message with the result (again, through the \emph{controller}) to the workflow execution engine.

The CWM is designed to cover complex curation tasks, which can potentially include large files. Since we want to avoid such larger files to use and thereby block resources for other processes, we implemented a priority feature in RabbitMQ queues. We reserve high priority processes for smaller documents and/or processes that take place in (semi-)real-time, while larger documents or more complex tasks can use normal/low priority for offline processing.

\subsection{Information Exchange Format}
\label{sec:architecture:informationformat}

Since interoperability is a key feature of the CWM, we must settle on a shared annotation format which all (or at least most\footnote{This is, first and foremost, relevant if tasks are relying on output of upstream tasks, or their output is input to downstream tasks.}) micro-services can work with and further augment in case of pipeline processing. Instead of defining our own format for this, we use the NLP Interchange Format (NIF) \cite{hellmann2013}. NIF includes an ontology that defines the way in which documents are annotated, with strong roots in the Linked Data paradigm. This allows for easy referencing of external knowledge bases (such as Wikidata) in the annotations on a document. NIF can be serialised in XML-like (RDF-XML), JSON-like (JSON-LD) or N3/turtle (RDF triple) formats. This serialised format is what is communicated as input or output for specific services. An example NIF (turtle) document with annotated named entities is shown in Listing~\ref{cod:nifexample} in the Appendix.

\subsection{Access Control}
\label{sec:architecture:restapi}

Access control for the various API endpoints is defined by the corresponding module, which specifies which operations are allowed for the endpoints of the different components, i.\,e., how a workflow is modified.

\begin{figure}[ht]
  \centering
  \includegraphics[width=\linewidth]{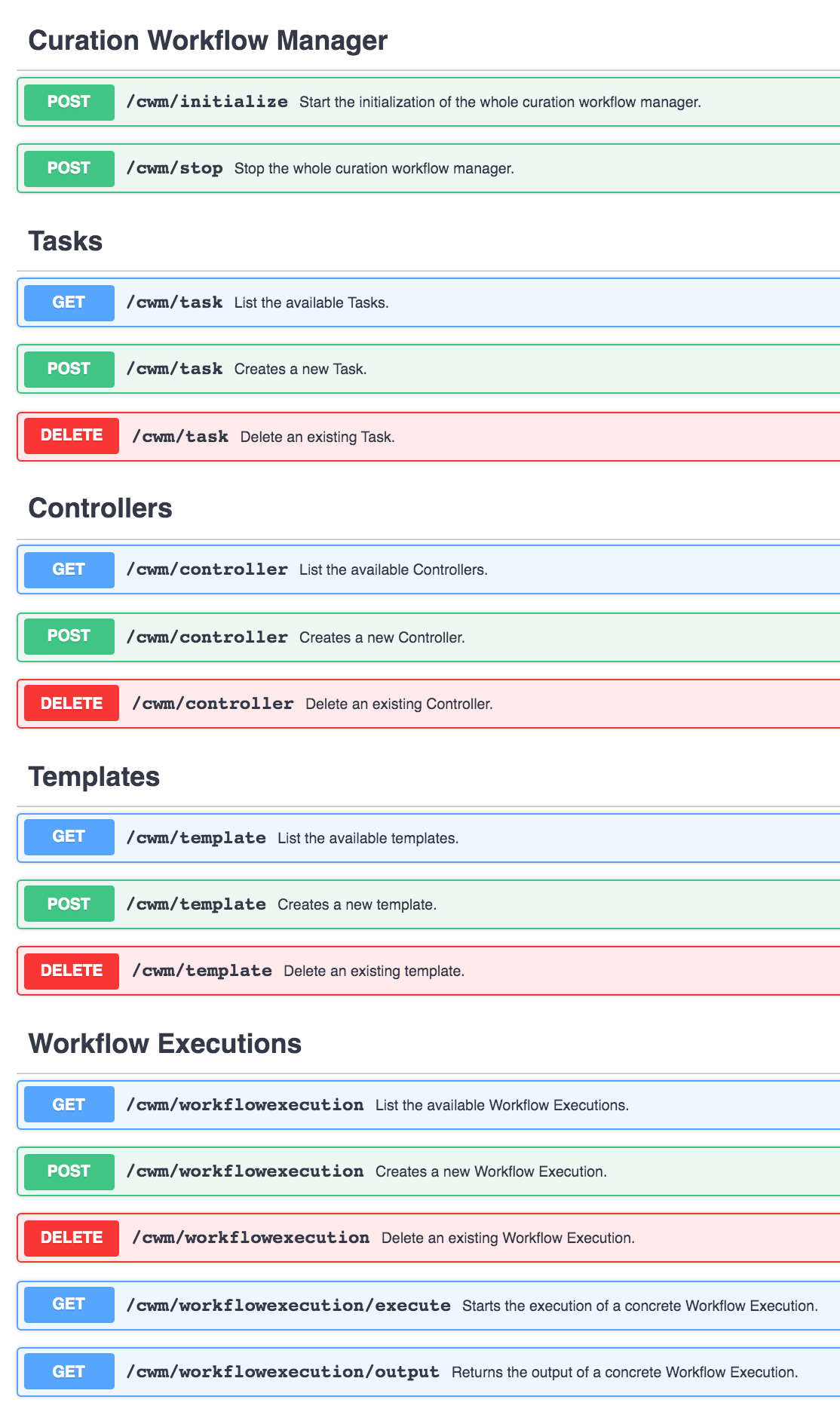}
  \caption{REST APIs}
  \label{fig:restapis}
\end{figure}

This module defines 12 methods that allow a user to (i) initialize and stop the CWM, (ii) view, create, modify and delete elements necessary to define workflows (i.\,e., \emph{tasks}, \emph{controllers}, \emph{templates} and \emph{workflow executions}, (iii) execute a specific workflow, and (iv) obtain the result of a workflow. An overview is provided in Figure~\ref{fig:restapis}.

In addition to the above mentioned functionalities, this module also handles security by allowing only users included in a pre-defined list to access the functionalities listed in Figure~\ref{fig:restapis}. We are currently working on more detailed user management by implementing user profiles, allowing certain users to access certain procedures only. This improvement will be included in a future version of the workflow manager.

\section{Conclusions and Future Work}
\label{sec:conclusionsfuturework}

We present an approach of connecting services and tools developed on different platforms and environments, in order to make them work together by means of a Curation Workflow Manager. The tool is built around the key principles of \emph{generality}, \emph{flexibility}, \emph{scalability} and \emph{efficiency}. It allows the combination of different tools, i.\,e., containerised micro-services, in the wider area of NLP, Information Retrieval, Question Answering, and Knowledge Management (triple stores) and uses a shared annotation format (NIF) throughout, addressing, respectively, the \emph{generality} and \emph{flexibility} principles. Our main motivation for developing the workflow manager, which comes with its own JSON-based definition language, was to address -- under the umbrella of a larger Curation Technology platform -- interoperability challenges and hardware-based resource-sharing and -handling issues in one go, addressing, respectively, the \emph{scalability} and \emph{efficiency} principles. 

The CWM is meant to process large documents, but is, as of now, restricted to text documents. As part of future work, we will also include the processing of multimedia files (images, audio, video). The curation workflow manager's design will be revised and extended accordingly. Furthermore, we plan to evaluate the workflow manager in a real-world use case provided by one of the partners in the QURATOR project. Additionally, we plan to integrate the CWM in the ELG platform in the medium to long term \cite{elg2020,labropoulou2020l,rehm2020n}. We currently work on extensions to the workflow definition language; its next iteration will be compliant with the standardised Business Process Model and Notation, increasing the sustainability and adaptability of our approach. Finally, we are currently considering the development of a visual editor (i.\,e., a GUI) to define and modify workflows, inspired by the GUI offered by Camunda\footnote{\url{https://camunda.com/products/modeler/}}.

% FIXME: Is it okay to leave this sentence in?
The source code of the Curation Workflow Manager is available on Gitlab.\footnote{\url{https://gitlab.com/qurator-platform/dfki/curationworkflowmanager}}

\section*{Acknowledgements}

% FIXME: Added ELG and Lynx. Okay?
The work presented in this paper has received funding from the German Federal Ministry of Education and Research (BMBF) through the project QURATOR (Wachstumskern no.~03WKDA1A) as well as from the European Union's Horizon 2020 research and innovation programme under grant agreements no.~825627 (European Language Grid) and no.~780602 (Lynx).

% \nocite{*}
\section{Bibliographical References}
\label{main:ref}

\bibliographystyle{./lrec}
\bibliography{./lrec2020}

\clearpage
\onecolumn

\section*{Appendix}

\begin{lstlisting}[backgroundcolor=\color{background},basicstyle=\ttfamily,showstringspaces=false,language=XML,frame=single,captionpos=b,caption=An example NIF document.,label=cod:nifexample]
<http://dkt.dfki.de/documents/#char=0,25>
 a                      nif:RFC5147String, nif:String, nif:Context ;
 nif:beginIndex         "0"^^xsd:nonNegativeInteger ;
 nif:endIndex           "25"^^xsd:nonNegativeInteger ;
 nif:isString           "Monteux was born in Paris"^^xsd:string .

<http://dkt.dfki.de/documents/#char=20,25>
 a                      nif:RFC5147String, nif:String ;
 nif:anchorOf           "Paris"^^xsd:string ;
 nif:beginIndex         "20"^^xsd:nonNegativeInteger ;
 nif:endIndex           "25"^^xsd:nonNegativeInteger ;
 nif:entity             <http://dkt.dfki.de/ontologies/nif#LOC> ;
 nif:referenceContext   <http://dkt.dfki.de/documents/#char=0,25> ;
 itsrdf:taIdentRef      <http://www.geonames.org/2988507> .

<http://dkt.dfki.de/documents/#char=0,7>
 a                      nif:RFC5147String, nif:String ;
 nif:anchorOf           "Monteux"^^xsd:string ;
 nif:beginIndex         "0"^^xsd:nonNegativeInteger ;
 nif:endIndex           "7"^^xsd:nonNegativeInteger ;
 nif:entity             <http://dkt.dfki.de/ontologies/nif#PER> ;
 nif:referenceContext   <http://dkt.dfki.de/documents/#char=0,25> ;
 itsrdf:taIdentRef      <http://d-nb.info/gnd/122700198> .
\end{lstlisting}

\end{document}